\documentclass[letterpaper, 10pt, conference]{ieeeconf}
\IEEEoverridecommandlockouts
\usepackage{cite}               
\usepackage{amsmath,amssymb,amsfonts}
\usepackage{mathtools}
\usepackage{algorithmic}
\usepackage{graphicx}
\usepackage{pgfplots}\pgfplotsset{compat=1.18}
\usepackage{booktabs}
\usepackage{array}
\usepackage{multirow}
\usepackage{subcaption}
\usepackage{xcolor}
\usepackage{siunitx}
\usepackage[hidelinks]{hyperref}
\usepackage{needspace}          

\newcommand{\GC}{GC}
\newcommand{\OA}{OA}
\newcommand{\SAS}{SAS}                    
\newcommand{\cmark}{\checkmark}
\newcommand{\xmark}{\ensuremath{\times}}

\begin{document}

\title{Mapless, Stabilizing, and Anti-Stuck Flipper Control\\ for Articulated Tracked Robots}

\author{Valent\'yn {\v C}\'ihala, Pavel Sarga, Martin Pecka, Tom\'a{\v s} Svoboda and Karel Zimmermann
\thanks{All authors are with Czech Technical University in Prague, Faculty of Electrical Engineering, Department of Cybernetics. Corresponding author V. {\v C}\'ihala:  cihalval@fel.cvut.cz.}%
}

\maketitle

\begin{abstract}
Nearly every autonomous or semi-autonomous flipper controller assumes access
to an \emph{elevation map} of the terrain under and in front of the robot. The terrain beneath the chassis is occluded in the sensors, so the map is remembered through odometry integration, whose drift corrupts it. We present
\emph{Stabilizing Anti-Stuck} (\SAS),
a mapless controller that combines a stabilization term for body roll
and pitch with an anti-stuck term that frees the robot when it gets stuck. As input, \SAS{} uses the IMU, the flipper joint angles, and a motion cue that only
tells whether the robot is moving. The cue is computed over a short window and never
accumulated, so \SAS{} does not inherit the odometry-integration drift that corrupts the
map. We deploy \SAS{} on a tracked robot with four flippers and evaluate it against manual
teleoperation and prior flipper controllers in a large comparison, both in
simulation and on a real obstacle. In simulation and on the real robot, \SAS{} achieves traversal quality comparable to state-of-the-art
map-based controllers. On a log hidden in tall grass, invisible to the LiDAR, the best map-based controller gets stuck
while \SAS{} traverses the hidden obstacle without a problem.
\end{abstract}

\section{Introduction}
Tracked and wheeled robots cannot ascend an obstacle taller than their leading wheel.
\emph{Articulated} tracked robots address higher obstacles with flippers,
independently actuated sub-tracks that reshape the robot's morphology on the fly: they
can be used to reach beyond the leading wheel to ascend what tracked and wheeled robots cannot, and to
free the robot when it gets \emph{stuck}~\cite{yuan2019,chen2023,xu2024}.

\begin{figure}[t]
  \centering
  \begin{tikzpicture}[inner sep=0, outer sep=0]
    \node[anchor=north west] (ra) at (0,0)
      {\includegraphics[width=\linewidth]{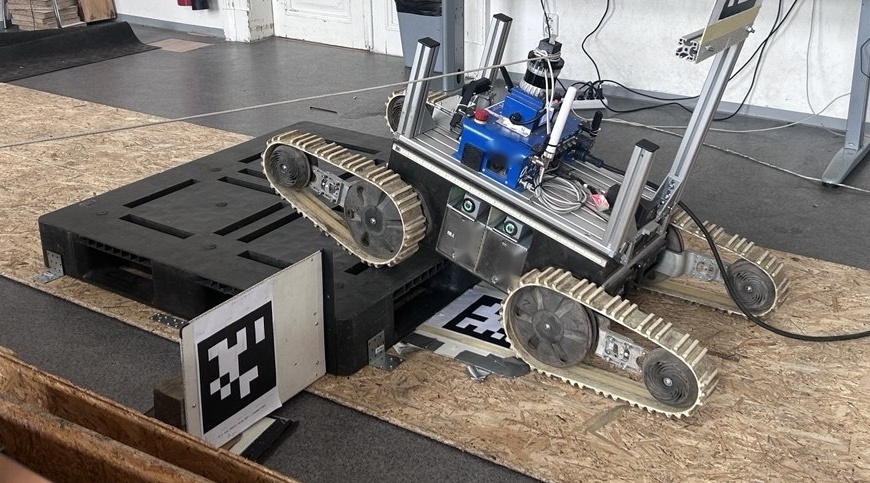}};
    \node[anchor=north west] (rb) at (ra.south west)
      {\includegraphics[width=\linewidth]{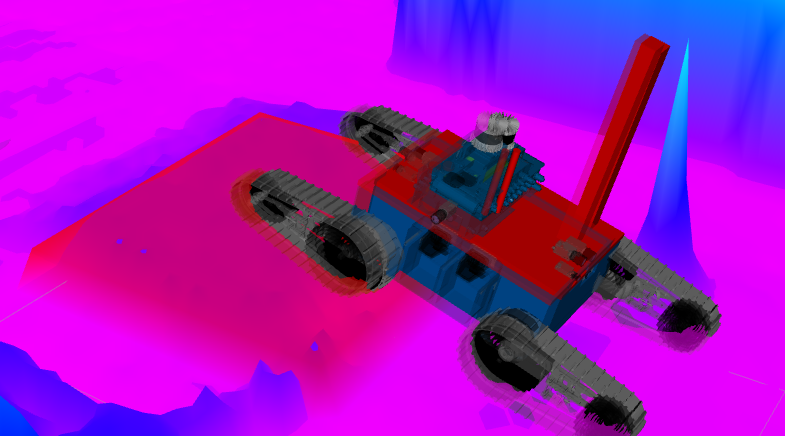}};
    \node[anchor=north west, inner sep=1.5pt, fill=white, fill opacity=0.75,
          text opacity=1, font=\bfseries\small] at (ra.north west) {(a)};
    \node[anchor=north west, inner sep=1.5pt, fill=white, fill opacity=0.75,
          text opacity=1, font=\bfseries\small] at (rb.north west) {(b)};
  \end{tikzpicture}
  \caption{\textbf{The onboard and ground-truth map on a pallet.} (a) The robot climbing a
  pallet obstacle. (b) The same pose, rendered from the data the robot observed there: the
  onboard map (blue) against the handcrafted ground-truth
  map positioned by AprilTags (transparent red), with the ground-truth robot pose drawn
  transparent. The offset between the two surfaces, and between the two bodies, is the map and localization error that the perception stack introduces. The spike behind the robot in the onboard map is a reflected, invalid LiDAR return.}
  \label{fig:robot}
\end{figure}

\textbf{Prior flipper controllers assume an elevation map.}
Across the literature, semi-autonomous and autonomous flipper controllers share a
precondition: they consume an \emph{elevation map} of the terrain under and in front of the
robot as their dominant
input~\cite{chen2023,oehler2024,yuan2019,xu2024,pan2023icm,pan2023atd3qn,ctrac2025,azayev2022}.
Fully manual teleoperation needs no map because the operator sees the terrain from a
third-person view but must drive all four flippers by hand. 

\textbf{On a real robot, the map is a dead-reckoning artifact.}
Where the estimate of that terrain drifts due to body occlusion, the commanded flipper
angles are computed for incorrect terrain. Onboard sensors
observe what is ahead, but the body occludes their view of the terrain beneath it.
Dedicated underbody sensors could measure it, but they are rarely fitted and have very
low resolution. That terrain can therefore mainly be \emph{remembered}: observed earlier, accumulated into an elevation
map, and carried beneath the robot as it advances by
\emph{integrating wheel or LiDAR odometry}~\cite{fankhauser2018}. Every stage of this
pipeline is a dead-reckoning process: point-cloud deskewing warps each scan by the estimated
sensor trajectory, and elevation mapping registers successive scans through the
estimated motion~\cite{loam2014}. On the loose, deformable,
step-like terrain where flippers are needed, odometry is least reliable: tracks slip, and
the chassis pitches and drops abruptly, saturating the IMU; therefore, inertial integration cannot
recover the estimate either~\cite{woodman2007imu}. Reflected, invalid LiDAR returns add obstacles that do
not exist, such as the spike behind the robot in Fig.~\ref{fig:robot}.

\textbf{The hardest case is a hidden obstacle.}
When a rigid obstacle is \emph{hidden} (Fig.~\ref{fig:grass}), optical sensors, such as LiDAR or cameras, return the
compliant canopy above it, while the load-bearing obstacle beneath remains invisible until the
chassis stalls on it. Neither a dedicated underbody sensor, which is blinded by the same vegetation, nor foliage-penetrating radar, which is bulky and low-resolution, resolves this in practice.

\textbf{Our contribution: control of the flippers without a map and without an accumulated
pose.}
We present \emph{Stabilizing Anti-Stuck} (\SAS), a reactive controller that requires no
elevation map. It has two terms: a
stabilization term that stabilizes the robot body regardless of the terrain, whether
climbing or descending an obstacle, and an anti-stuck term that frees the robot when it
gets stuck by driving the flippers down. To tell a stuck robot from a moving one, the
anti-stuck term needs a motion cue: a thresholded answer to whether the robot is
moving, which requires neither an accurate nor a metric estimate of the motion. Because the
cue is computed over a short window and is never accumulated, \SAS{} never inherits the drift that corrupts the map.

\needspace{5\baselineskip}%
\textbf{Contributions:}
\nopagebreak
\begin{enumerate}
  \item We propose \SAS, a mapless flipper controller that needs only a
  thresholded motion cue, combining continuous roll and pitch stabilization with a reactive
  anti-stuck term under a safe directional-priority blend.
  \item We measure the degradation of the elevation map caused by body occlusion and
  odometry drift by evaluating map-based baselines under both a \emph{ground-truth} map and
  the \emph{onboard} map, in a large comparison in simulation of \SAS{} against \emph{twelve}
  map-based baselines and two manual baselines on a twelve-obstacle course; every controller
  drives the full course ten times, with the perception stack reset at each obstacle.
  \item We perform the large comparison in a real obstacle arena with AprilTag ground
  truth: on a four-track robot, each controller drives ten times over a pallet obstacle.
\end{enumerate}

\section{Related Work}\label{sec:related}
We group the flipper controllers into manual control and three families of autonomous
control: reactive and geometry-based, imitation learning, and reinforcement learning.
Table~\ref{tab:sensing} summarizes the compared controllers and what they
require.

\subsection{Manual control}
In direct teleoperation, the operator drives the flippers by hand~\cite{cihala2025}, or
switches between preset flipper modes~\cite{larochelle2013,azayev2022}. Manual control
needs no map: the operator observes the terrain from a third-person view, closes the control
loop by hand, and also stabilizes the robot.

\subsection{Reactive and geometry-based control}
These controllers are the most dependent on an explicit terrain model because they compute the flipper commands either
reactively~\cite{cihala2025} or by planning. Chen et al. combine dynamic programming with a
fast geometry-based pose predictor over the elevation
point cloud~\cite{chen2023}; Yuan et al. plan in the flipper configuration space
from known stair geometry with motion-capture ground truth~\cite{yuan2019}; Oehler
and von Stryk predict pose on a signed-distance field built from
LiDAR~\cite{oehler2024}; Xu et al. solve a receding-horizon hybrid trajectory
optimization over a simplified terrain sampled from a prebuilt 3D occupancy map with
FAST-LIO2 odometry~\cite{xu2024}; and most recently, Zhang et al. plan the path and flipper
maneuvers with RRT* over a contact-model pose predictor, using learned semantic
segmentation of the elevation map only as a perception front-end~\cite{zhang2025}.

\subsection{Imitation learning}
These controllers learn from demonstrations which low-level behavior to apply. The
adaptive-traversability line~\cite{zimmermann2014} classifies
terrain into discrete flipper modes from an elevation map. 
A learned state machine~\cite{azayev2022}
switches among seven hand-crafted flipper morphologies selected from an elevation map. In addition, each
state overlays a \emph{roll}-stabilization term and a stagnation-triggered
\emph{anti-stuck} term that lowers all four flippers to free the robot~\cite{azayev2022}.

\subsection{Reinforcement learning}
Pecka et al. learn a
safety-constrained policy (Constrained REPS) whose state is the body pitch and the terrain
height \SI{20}{\centi\metre} ahead of the robot, observed from an occupancy map built online from the LiDAR scans~\cite{pecka2016}. Discrete-action variants dominate the
flipper-specific literature: Pan et al. learn a dueling double-DQN over discrete
front/rear flipper increments~\cite{pan2023atd3qn} and extend it with an
intrinsic-curiosity module~\cite{pan2023icm}; both take the chassis pitch and the average
terrain heights of a few segments ahead of, below, and behind the robot. Their reward favors smooth pitch only, and getting stuck is a terminal
penalty. Mitriakov et al.
learn staircase negotiation with a policy-gradient method from step-edge geometry~\cite{mitriakov2021}, injecting stability through center-of-gravity and
pitch-rate reward shaping. Most recently, C-TRAC learns a contact-aware latent representation from noisy LiDAR and
drives an asymmetric-SAC policy that stabilizes both roll and pitch via a
normalized-energy-stability-margin reward~\cite{ctrac2025}. We train our RL baselines in
FTR-Bench~\cite{ftrbench2025}, an Isaac Lab framework that benchmarks standard RL
algorithms on a flipper robot.

\subsection{Attitude stabilization and mapless posture control}\label{sec:related-posture}
Flipper-based attitude stabilization has been studied before, but nearly always on top of
an elevation map: a roll-only stabilization term inside a map-based state
machine~\cite{azayev2022}, and C-TRAC stabilizing both roll and pitch from a reward computed
on the map-based simulator state~\cite{ctrac2025}. Rocha et al.~\cite{rocha2023} regulate
roll, pitch, and ground clearance of an articulated tracked vehicle from the IMU and the
flipper joints.

\subsection{Evaluating traversal quality}
To evaluate traversal quality, multiple metrics were proposed: IMU-based smoothness
measured on the robot's body, such as pitch swing and vertical
shock~\cite{chen2023}, body attitude, and chassis-to-ground clearance read from the
ground-truth map~\cite{cihala2025,zimmermann2014}.

\begin{table}[t]
\centering
\caption{\textbf{Compared controllers.} \emph{Map}: needs an onboard
map. \emph{Stab.}: attitude axes with an explicit stabilization objective (r roll,
p pitch). \emph{Anti-stuck}: frees itself when stuck. \emph{Code}: the authors' released
implementation or our reimplementation.}
\label{tab:sensing}
\scriptsize
\setlength{\tabcolsep}{1.3pt}
\begin{tabular}{@{}llcccc@{}}
\toprule
Controller & Approach & Code & Map & Stab. & \shortstack{Anti-\\stuck} \\
\midrule
\multicolumn{6}{@{}l}{\emph{Manual}} \\
Manual control~\cite{cihala2025} & direct teleoperation & -- & \xmark & -- & \xmark \\
Manual modes~\cite{azayev2022} & operator switches preset modes & -- & \xmark & -- & \xmark \\
\midrule
\multicolumn{6}{@{}l}{\emph{Reactive and geometry-based}} \\
{\v C}\'ihala \GC/\OA~\cite{cihala2025} & clearance + obstacle-aware & released & \cmark & -- & \xmark \\
Chen GFMP~\cite{chen2023} & pose prediction + dynamic prog. & reimpl. & \cmark & p & \xmark \\
Yuan~\cite{yuan2019} & configuration-space planning & reimpl. & \cmark & -- & \xmark \\
Oehler~\cite{oehler2024} & SDF pose prediction & released & \cmark & -- & \xmark \\
Xu HTO~\cite{xu2024} & hybrid trajectory optimization & reimpl. & \cmark & p & \xmark \\
Zhang~\cite{zhang2025} & contact-model pose prediction & reimpl. & \cmark & -- & \xmark \\
\midrule
\multicolumn{6}{@{}l}{\emph{Imitation learning}} \\
Azayev~\cite{azayev2022} & learned template state m. & released & \cmark & r & \cmark \\
\midrule
\multicolumn{6}{@{}l}{\emph{Reinforcement learning}} \\
Pecka~\cite{pecka2016} & safety-constrained REPS & released & \cmark & p & \xmark \\
Pan AT-D3QN~\cite{pan2023atd3qn} & dueling double-DQN & reimpl. & \cmark & p & \xmark \\
Pan ICM-D3QN~\cite{pan2023icm} & D3QN + intrinsic curiosity & reimpl. & \cmark & p & \xmark \\
Mitriakov~\cite{mitriakov2021} & policy-gradient stair climbing & released & \cmark & p & \xmark \\
C-TRAC~\cite{ctrac2025} & contact-aware asym. SAC & reimpl. & \cmark & r\&p & \xmark \\

\midrule
\textbf{\SAS{} (ours)} & stabilization + anti-stuck & ours & \xmark & r\&p & \cmark \\
\bottomrule
\end{tabular}
\end{table}

\section{Method}\label{sec:method}
At every control step, $\Delta t=\SI{0.05}{\second}$ i.e. $f_{control}=20 ~\text{Hz}$, \SAS{} reads the gravity-referenced body roll $\phi$ and pitch $\theta$ from
the IMU, the four flipper joint angles $q_i$, the commanded body velocity $v_\text{cmd}$,
and a scalar motion cue $m$ (Sec.~\ref{sec:method-antistuck}).

The output of \SAS{} is an angular \emph{velocity} command $v_i$ for each flipper
$i\in\{\mathrm{FL},\mathrm{FR},\mathrm{RL},\mathrm{RR}\}$. The command $v_i$ is composed of two terms: the
stabilization term $v^{\mathrm{s}}_i$, which stabilizes body roll and pitch, and the
anti-stuck term $v^{\mathrm{a}}_i$, which frees the robot when it is \emph{stuck}.

\subsection{Attitude stabilization}\label{sec:method-stab}
Attitude stabilization drives the body roll $\phi$ and pitch $\theta$ to the zero set points
$\phi_0=\theta_0=0$, which can later be set to a target attitude: from the errors $e_\phi=\phi_0-\phi$ and $e_\theta=\theta_0-\theta$, we form the proportional-integral roll command $u_\phi$ and
pitch command $u_\theta$.
\begin{equation}
  u_\phi = K^\phi_{\mathrm P}\,e_\phi + K^\phi_{\mathrm I}\, I_\phi, \qquad
  u_\theta = K^\theta_{\mathrm P}\,e_\theta + K^\theta_{\mathrm I}\, I_\theta,
  \label{eq:pi}
\end{equation}
where $I_\phi$ and $I_\theta$ are the anti-windup-limited integrals,
$I_{\phi,\theta}\leftarrow\operatorname{sat}(I_{\phi,\theta}+e_{\phi,\theta}\Delta t,\ I_{\max})$.
The two commands are distributed to the four flippers as a front/rear differential for
pitch and a left/right differential for roll:
\begin{equation}
  \begin{bmatrix} v^{\mathrm{s}}_{\mathrm{FL}}\\ v^{\mathrm{s}}_{\mathrm{FR}}\\ v^{\mathrm{s}}_{\mathrm{RL}}\\ v^{\mathrm{s}}_{\mathrm{RR}} \end{bmatrix}
  =
  \begin{bmatrix} -1 & +1\\ -1 & -1\\ +1 & +1\\ +1 & -1 \end{bmatrix}
  \begin{bmatrix} u_\theta\\ u_\phi \end{bmatrix}.
  \label{eq:stabmap}
\end{equation}

\subsection{Reactive anti-stuck}\label{sec:method-antistuck}
The anti-stuck term follows~\cite{azayev2022}, the one prior method that couples an anti-stuck term with attitude stabilization: when the robot is stuck, it drives the
flippers to lift the body off the ground, freeing the robot. The \emph{dig} condition is defined as
\begin{equation}
    \text{dig} \coloneqq \big(|v_\text{cmd}|>\tau_\text{move}\big)\ \wedge\
  \big(m<\tau_\text{stuck}\big),
  \label{eq:stuck}
\end{equation}
where $\tau_\text{move}=\SI{0.1}{\metre\per\second}$ is the commanded velocity threshold and $\tau_\text{stuck}=\SI{2}{\milli\metre}$ is the threshold of the motion cue $m$.
The motion cue $m$ is the standard deviation of the horizontal position norm
$\sqrt{x^2+y^2}$ of the robot from LiDAR-ICP
odometry~\cite{libpointmatcher} over a sliding \SI{0.5}{\second} window. The cue only has
to answer whether the robot is moving: it is thresholded, so it needs neither an accurate
nor a metric estimate of the motion, and any scale-free motion signal would serve. It is
computed over the window and never accumulated, so it carries none of the drift of an
integrated pose. Unlike track odometry, LiDAR odometry is capable of detecting no movement if the tracks slip
on a stuck robot. When digging,
all four flippers are driven downward at the velocity
$\lambda_\text{dig}=\SI{2.0}{\radian\per\second}$ so they push against the terrain and lift
the body up; otherwise, they retract with the velocity
$\lambda_\text{retract}=\SI{0.8}{\radian\per\second}$:
\begin{equation}
  v^{\mathrm{a}}_i = \begin{cases} +\lambda_\text{dig} & \text{dig}\\ -\lambda_\text{retract} & \text{otherwise.}\end{cases}
  \label{eq:spread}
\end{equation}
Digging alone can lock up: if the flippers reach their lowest position limit without
freeing the robot, it would remain stuck indefinitely with its flippers fully down.
Therefore, a timer $T_\text{dig}=\SI{8}{\second}$ monitors the dig itself: after $T_\text{dig}$ of continuous digging without release, the anti-stuck term
is suspended for $T_\text{reset}=\SI{2.5}{\second}$ while the flippers are driven back up, after which digging is permitted again.

\subsection{Directional-priority blend and limits}\label{sec:method-blend}
The two terms are combined by the directional-priority blend: the $\mathrm{cap}$ operator
limits the anti-stuck term whenever it opposes the stabilization term on a flipper to a
fraction $\alpha\in[0,1]$ of the stabilization velocity, so their sum keeps the sign of the stabilization
term. The anti-stuck term can therefore reinforce or nullify the stabilization command but
never reverse its direction on any flipper:
\begin{equation}
  \mathrm{cap}(v^{\mathrm{a}}_i,v^{\mathrm{s}}_i)=
  \begin{cases}
        \operatorname{sat}\big(v^{\mathrm{a}}_i,\ \alpha |v^{\mathrm{s}}_i|\big)
      & \text{if } v^{\mathrm{s}}_i v^{\mathrm{a}}_i<0 \wedge |v^{\mathrm{s}}_i|>\beta,\\[2pt]
    v^{\mathrm{a}}_i & \text{otherwise,}
  \end{cases}
      \label{eq:cap}
\end{equation}
where $\operatorname{sat}(x,c)=\max(-c,\min(x,c))$. 
The blended command is saturated to
the velocity limit $v_{\max}=\SI{1.0}{\radian\per\second}$ of the flipper drives:
\begin{equation}
  v_i = \operatorname{sat}\big(\, v^{\mathrm{s}}_i + \mathrm{cap}(v^{\mathrm{a}}_i,\,v^{\mathrm{s}}_i),\ v_{\max}\big).
  \label{eq:overview}
\end{equation}

The fraction $\alpha$ sets how much the opposing anti-stuck term may weaken stabilization. When the condition $v^{\mathrm{s}}_i v^{\mathrm{a}}_i~<~0$ applies, the $\mathrm{cap}$ saturates the anti-stuck command to the fraction $\alpha |v^{\mathrm{s}}_i|$ of the stabilization velocity. If $\alpha=0$ and $v^{\mathrm{s}}_i v^{\mathrm{a}}_i~<~0$, then $\mathrm{cap}(v^{\mathrm{a}}_i,v^{\mathrm{s}}_i)=0$, the anti-stuck term yields completely, and equation~\eqref{eq:overview} simplifies to $v_i=\operatorname{sat}(v^{\mathrm{s}}_i, v_{\max})$. In the case of $\alpha=1$, the anti-stuck term can
reach $\mathrm{cap}(v^{\mathrm{a}}_i,v^{\mathrm{s}}_i)=-v^{\mathrm{s}}_i$; therefore, equation~\eqref{eq:overview} simplifies to $v_i~=~0$. This enforces that the anti-stuck command can nullify, but never reverse the stabilization command. We use $\alpha=0.5$, so a conflicting dig is allowed at half strength. The cap
applies only while $|v^{\mathrm{s}}_i|>\beta=\SI{0.05}{\radian\per\second}$, so that
attitude noise on a level robot cannot suppress the anti-stuck term; otherwise, the
anti-stuck command passes through unchanged.

\section{Experiments}\label{sec:experiments}
\subsection{Platform and onboard pipeline}
The experimental robot is an articulated \emph{four-track} robot whose four independently actuated flippers are also its only
driven tracks (Fig.~\ref{fig:robot}). The robot software backbone is ROS 2, which allows for the development of a distributed robotic system. We evaluate the flipper controllers both in simulation and on the
real platform. The onboard perception pipeline that the map-based baselines
consume is representative of current practice: an Ouster LiDAR and IMU feed point-cloud
deskewing and LiDAR-ICP odometry~\cite{norlabicpmapper,libpointmatcher}, which drive elevation
mapping~\cite{fankhauser2018} to produce a robot-centric
elevation map, which we call the \emph{onboard map}.

\begin{figure*}[t]
  \centering
  \includegraphics[width=\textwidth]{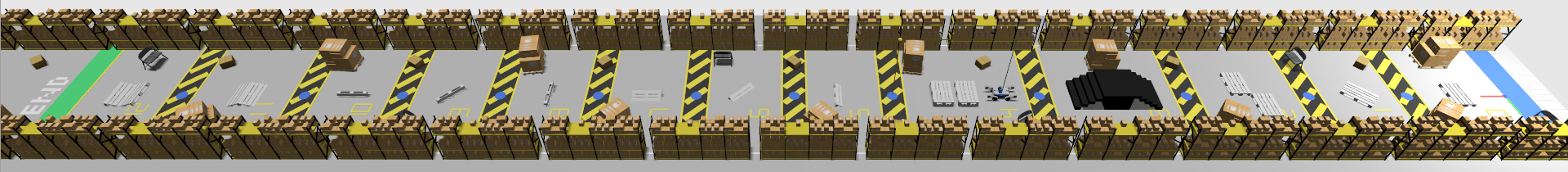}
  \caption{\textbf{Simulation arena.} The warehouse course: a straight aisle of twelve numbered obstacles of variable difficulty (pallets, stacked
  boxes, planks, and a staircase); the robot starts at the right. In the autonomous run (Sec.~\ref{sec:experiments}) it is driven at a constant
  \SI{0.6}{\metre\per\second} through every obstacle in turn, with a \SI{1}{\metre} settling strip between obstacles.}
  \label{fig:obstacles}
\end{figure*}

\paragraph{Simulation comparison} We simulate in Gazebo Ionic~\cite{gazebo2004} with the DART physics engine. Unlike
typical simulators that approximate tracks with multiple wheels, we employ the track
simulation of~\cite{pecka2017tracks}, which models conveyor-belt behavior. The sensors match the real payload: a simulated Ouster OS0
($1024\times64$ rays at \SI{10}{\hertz}, \SIrange{0.1}{50}{\metre}, \SI{1}{\centi\metre}
Gaussian range noise) and a \SI{50}{\hertz} IMU with Gaussian noise and a slowly varying
bias. The onboard-map condition runs the robot's ICP and elevation-mapping nodes on these
simulated ROS 2 topics. Ground-truth pose and map for evaluation are taken from the
simulator. The arena is a twelve-obstacle warehouse course (Fig.~\ref{fig:obstacles}): a straight aisle lined with twelve numbered obstacles of
variable difficulty, including pallets, stacked boxes, planks, and a staircase.

\paragraph{Real robot comparison} The real arena repeats the simulation experiment on one pallet obstacle between
two reset zones (Fig.~\ref{fig:robot}). AprilTag markers on the robot provide absolute localization in the handcrafted
ground-truth map.

\paragraph{Autonomous run and \SAS{} override} In all reported controller runs, the robot is driven and steered autonomously, ensuring that the comparison is not biased by an operator's control; in the two manual references, which set the reference level, the operator drives the robot and all four flippers by hand from a third-person view. The autonomous run commands the robot forward at a constant velocity
\SI{0.6}{\metre\per\second} in simulation and, for safety, \SI{0.25}{\metre\per\second}
on the real robot. A controller that stalls on an obstacle is released by the \emph{\SAS{} override}, which
hands the flippers to \SAS{} anti-stuck dig mode until the robot moves again.

\subsection{Baselines}
We evaluate \SAS{} together with two manual references: \textbf{Manual control}, an expert operator driving all four flippers by hand from a
third-person view of the robot, and \textbf{Manual modes}, the same operator switching between preset flipper modes by hand~\cite{azayev2022}, and with every map-based controller cataloged in Table~\ref{tab:sensing}. 

Where the authors released code, we run their implementation; where they did not, we use our reimplementation from the paper.

The geometry-based controllers were adapted to the four-track morphology as follows. Chen's
GFMP~\cite{chen2023} keeps the exact two-contact solver and waypoint dynamic programming but admits only flipper-supported contact cases,
since the chassis is passive, adds an absolute-pitch levelness cost, and uses the
four-track morphology and the onboard-map profile. Yuan's configuration-space
planner~\cite{yuan2019} was reimplemented for the four-track morphology; it estimates the
step distance and height at every control step from the onboard map, replans the flipper trajectory continuously, and holds the flippers where its model is undefined. Oehler and von Stryk's pose predictor~\cite{oehler2024}, given a four-flipper model with a passive chassis, gains the selection stage it lacks: a
43-candidate sweep commanding the highest-stability-margin configuration, holding when
no configuration is stable. Xu's HTO~\cite{xu2024} assumes a driven chassis that carries the step edge; with a passive chassis, the chassis-contact regimes of its hybrid model are replaced on ascents by a double-prop
flipper bridge with enforced chassis clearance, plus rear-release and
descent-preparation schedules. Zhang's planner~\cite{zhang2025} is reproduced only as its
contact-model pose prediction and action cost, solved at a virtual station
\SI{0.5}{\metre} ahead; chassis-supported poses are forbidden on the passive chassis. The mapless posture controller of Rocha
et al.~\cite{rocha2023} is not evaluated, since it is not designed for obstacle traversal.

The learned controllers were adapted as follows. Pan's
AT-D3QN~\cite{pan2023atd3qn}: the flipper reward $R_{\text{flipper}}$ is mirrored onto the
rear flippers, and the target pitch is referenced to the ground rather than to the chassis,
since on a four-track robot the rear flippers are part of the drive train and the ride
height is a free variable; the contact reward $R_{\text{contact}}$ of ICM-D3QN is added and
redefined around flipper-pair grounding and chassis clearance. Pan's
ICM-D3QN~\cite{pan2023icm}: the same three changes, sharing the code with AT-D3QN;
otherwise, as published. C-TRAC~\cite{ctrac2025}: reward terms are added that incentivize
movement and prevent convergence to a fixed pose; trained at a fixed forward velocity.
Mitriakov~\cite{mitriakov2021}: the 2-DoF arm is dropped, so the center-of-gravity penalty
reduces to a chassis-tilt proxy; one policy covers both driving directions, and the rule
that the trailing flipper never goes below neutral is inverted, since on a four-track robot
that flipper carries a load; trained at a fixed forward velocity. Pecka~\cite{pecka2016}: the
sign convention follows the authors' reference implementation rather than the paper, and
safety is read from the same rollout.

\subsection{Ground-truth map for a fair baseline}
The map-based baselines depend on the onboard map, so scoring them under that map alone
would mix the quality of the controller with the quality of its perception. We therefore
run each of them twice: with the onboard map and localization, and with a
\emph{ground-truth} map that is free of odometry noise. The difference between the two scores is the baseline's
sensitivity to onboard mapping.

\subsection{Metrics}
Following~\cite{cihala2025} we report \textbf{traversal quality} (TQ), the mean of three
normalized and distance-binned channel scores of a traversal, shock $q_s$, body tilt
$q_t$, and ground clearance $q_c$,
\begin{equation}
  \mathrm{TQ}=\tfrac{1}{3}\,(q_s+q_t+q_c)\in(0,1].
  \label{eq:tq}
\end{equation}
We also report the percentage of the obstacle distance for which the \SAS{} override dig drove the flippers.
Traversal success and traversal time are not informative in this setting: the override
releases every stall, so every obstacle is completed, and the drive node drives at a constant speed; therefore, the traversal time differs between controllers only by the time spent in the stall state before the override activates.
\label{sec:metrics-block}%

\noindent\textbf{Shock} measures the impacts transmitted to the body, which stress
the robot and its payload. It is the acceleration magnitude $s=|a|$ from the IMU, normalized with the scale $s_{\max}$, the
largest acceleration the platform records. On the real robot $s_{\max}$ is the highest measured
sample of the comparison, \SI{57.5}{\metre\per\second\squared}. In simulation, the rigid contacts of DART produce single-step spikes, so $s_{\max}$ is the highest measured \SI{0.1}{\second} mean of
$s$ over all runs, \SI{376}{\metre\per\second\squared}.

\noindent\textbf{Tilt} measures how far the body departs from the gravity-aligned
horizontal; it is what the attitude stabilization regulates, and it measures how safely the controller rides. It is the angle
$t$ between the body $z$ axis and the vertical, from a level pose, normalized with
$t_{\max}=\ang{50}$.

\noindent\textbf{Ground clearance} measures whether the chassis is kept off the terrain. We measure it as the
closest ground point of a $10\times8$ grid with \SI{5}{\centi\metre}
cells under the chassis, covering its $50\times\SI{40}{\centi\metre}$ footprint, read from the ground-truth map. The four-track morphology leaves a \SI{7.3}{\centi\metre} gap under the chassis on flat
ground with level flippers, which the score takes as its reference. Rewarding raw clearance would favor
baselines that freeze their flippers and coast in a high pose.

\noindent\textbf{Normalization.} The three channels are physical quantities in different
units, and each has to become a score in $(0,1]$ where $1$ is the best, so that they can be
averaged in~\eqref{eq:tq} and compared across controllers. Shock and tilt are unbounded above, so a non-negative sample $x$ is
mapped by the sigmoid of~\cite{cihala2025},
\begin{equation}
  N(x,\hat{x}) = 2 - \frac{2}{1 + e^{-x/\hat{x}}},
  \label{eq:norm}
\end{equation}
against a reference $\hat{x}$: $N=1$ at $x=0$, $N\approx0.54$ when the sample reaches
$\hat{x}$, and $N\to0$ for large samples. Each channel is scored as $q=N(x,x_{\max}/2)$ with
its own scale $x_{\max}$, so the score is close to linear for small samples and saturates
smoothly; a single extreme sample cannot dominate the average. 

\noindent\textbf{Distance binning} is used rather than time binning 
as each controller takes a different amount of time to traverse the same obstacle. The measured data are divided into bins of \SI{0.25}{\metre}
along the driving direction of an obstacle; the worst per-sample score in each bin is kept, and the final
channel score is the mean of these bin values. 

Averaging the three channels in~\eqref{eq:tq} assumes that they carry different
information, so they should not correlate strongly; their pairwise correlations
over the traversed obstacles (Table~\ref{tab:spearman}) are at most $0.39$, which suggests that none of our evaluation channels are redundant.

\begin{table}[t]
\centering
\caption{\textbf{Correlation of the channel scores} over the traversed simulated
obstacles and real crossings.}
\label{tab:spearman}
\footnotesize
\begin{tabular}{@{}lcc@{}}
\toprule
Channel pair & Simulation & Real \\
\midrule
Shock -- clearance & $0.26$ & $0.06$ \\
Shock -- tilt & $0.28$ & $0.39$ \\
Clearance -- tilt & $0.39$ & $-0.03$ \\
\bottomrule
\end{tabular}
\end{table}

\begin{table*}[t]
\centering
\caption{\textbf{Traversal quality in simulation and on the real pallet.} Simulation: mean over
ten runs of the twelve-obstacle course. Real: one ride of ten crossings per controller and map, mean over
the crossings. All channels are scored against ground
truth, with $s_{\max}$ \SI{376}{\metre\per\second\squared} in simulation and
\SI{57.5}{\metre\per\second\squared} on the real robot. Each cell gives the value under the ground-truth map / the onboard map. Shock,
clearance and tilt are the three channel scores of~\eqref{eq:tq}. \emph{Override}: percentage of
distance an anti-stuck dig drove the flippers, by the \SAS{} override for the baselines and by its own term for \SAS{}. Manual rides are hand-driven from a third-person view. Per column and map condition, the best controllers are in bold and the second best underlined (ties share the mark).}
\label{tab:results}
\scriptsize
\setlength{\tabcolsep}{1.5pt}
\begin{tabular*}{\textwidth}{@{\extracolsep{\fill}}lc!{\vrule width 0.4pt}ccc!{\color{black!35}\vrule width 0.4pt}c!{\vrule width 0.4pt}c!{\vrule width 0.4pt}ccc!{\color{black!35}\vrule width 0.4pt}c@{}}
\toprule
 & \multicolumn{5}{c}{Simulation, twelve obstacles} & \multicolumn{5}{c}{Real pallet, one ride of ten crossings} \\
\cmidrule(lr){2-6}\cmidrule(l){7-11}
Controller & TQ $\uparrow$ & Shock $\uparrow$ & Clearance $\uparrow$ & Tilt $\uparrow$ & Override [\%] $\downarrow$ & TQ $\uparrow$ & Shock $\uparrow$ & Clearance $\uparrow$ & Tilt $\uparrow$ & Override [\%] $\downarrow$ \\
 & \multicolumn{10}{c}{each cell: ground-truth map / onboard map} \\
\midrule
\multicolumn{11}{@{}l}{\emph{Manual}} \\
Manual control~\cite{cihala2025} & -- / 0.888 & -- / 0.922 & -- / 0.866 & -- / 0.875 & -- & -- / 0.810 & -- / 0.775 & -- / 0.714 & -- / 0.942 & -- \\
Manual modes~\cite{azayev2022} & -- / 0.846 & -- / 0.916 & -- / 0.788 & -- / 0.834 & -- & -- / 0.810 & -- / 0.766 & -- / 0.816 & -- / 0.848 & -- \\
\midrule
\multicolumn{11}{@{}l}{\emph{Reactive and geometry-based}} \\
{\v C}\'ihala \GC/\OA~\cite{cihala2025} & 0.820 / 0.801 & \textbf{0.918} / \textbf{0.916} & \textbf{0.745} / \textbf{0.703} & 0.798 / 0.783 & \underline{0.2} / \textbf{0.0} & 0.695 / 0.657 & 0.780 / \underline{0.794} & \underline{0.528} / 0.323 & 0.777 / 0.854 & \textbf{0.0} / \textbf{0.0} \\
Chen GFMP~\cite{chen2023} & \underline{0.830} / \underline{0.810} & 0.909 / 0.905 & 0.726 / 0.667 & \underline{0.854} / \underline{0.857} & 9.0 / 9.2 & 0.660 / 0.661 & \underline{0.792} / 0.789 & 0.301 / 0.299 & \textbf{0.886} / \textbf{0.895} & 3.1 / 3.0 \\
Yuan~\cite{yuan2019} & 0.828 / 0.803 & \underline{0.912} / \underline{0.911} & \underline{0.736} / 0.667 & 0.836 / 0.832 & 11.3 / 8.8 & 0.710 / \textbf{0.721} & 0.776 / 0.783 & 0.471 / \underline{0.497} & \underline{0.883} / 0.883 & \underline{1.3} / 1.2 \\
Oehler~\cite{oehler2024} & 0.825 / 0.797 & 0.904 / 0.902 & 0.733 / 0.658 & 0.837 / 0.832 & 10.2 / 7.6 & \textbf{0.713} / 0.651 & 0.785 / 0.787 & 0.500 / 0.312 & 0.854 / 0.854 & 2.8 / 1.5 \\
Xu HTO~\cite{xu2024} & 0.807 / 0.794 & 0.899 / 0.898 & 0.690 / 0.652 & 0.833 / 0.833 & 9.5 / 9.8 & 0.685 / 0.677 & 0.779 / 0.776 & 0.404 / 0.387 & 0.872 / 0.868 & 3.5 / 3.3 \\
Zhang~\cite{zhang2025} & 0.811 / 0.797 & 0.911 / 0.904 & 0.694 / 0.654 & 0.828 / 0.832 & 11.6 / 14.4 & 0.689 / 0.692 & 0.779 / 0.780 & 0.427 / 0.441 & 0.861 / 0.856 & 3.4 / 3.1 \\
\midrule
\multicolumn{11}{@{}l}{\emph{Imitation learning}} \\
Azayev~\cite{azayev2022} & 0.797 / 0.781 & 0.865 / 0.858 & 0.698 / 0.655 & 0.828 / 0.831 & 7.6 / \underline{7.1} & 0.662 / 0.695 & 0.775 / 0.775 & 0.347 / 0.424 & 0.864 / 0.885 & \textbf{0.0} / \textbf{0.0} \\
\midrule
\multicolumn{11}{@{}l}{\emph{Reinforcement learning}} \\
Pecka~\cite{pecka2016} & 0.808 / 0.792 & 0.896 / 0.888 & 0.700 / 0.660 & 0.827 / 0.829 & 10.0 / 10.3 & 0.642 / 0.640 & 0.765 / 0.764 & 0.411 / 0.398 & 0.751 / 0.757 & \textbf{0.0} / \textbf{0.0} \\
Pan AT-D3QN~\cite{pan2023atd3qn} & 0.807 / 0.792 & 0.904 / 0.898 & 0.694 / 0.652 & 0.823 / 0.826 & 10.6 / 12.1 & 0.642 / 0.644 & 0.773 / 0.770 & 0.307 / 0.313 & 0.845 / 0.849 & 3.5 / \underline{0.4} \\
Pan ICM-D3QN~\cite{pan2023icm} & 0.801 / 0.793 & 0.908 / 0.903 & 0.677 / 0.645 & 0.819 / 0.831 & 10.2 / 12.2 & 0.652 / 0.666 & 0.774 / 0.771 & 0.318 / 0.391 & 0.864 / 0.836 & 4.7 / 6.1 \\
Mitriakov~\cite{mitriakov2021} & 0.807 / 0.783 & 0.885 / 0.876 & 0.716 / 0.653 & 0.821 / 0.820 & 10.0 / 8.6 & \underline{0.712} / 0.635 & 0.781 / 0.776 & 0.501 / 0.323 & 0.855 / 0.806 & \textbf{0.0} / 2.3 \\
C-TRAC~\cite{ctrac2025} & 0.821 / 0.752 & 0.887 / 0.895 & 0.722 / 0.608 & \underline{0.854} / 0.754 & \textbf{0.0} / 7.5 & 0.665 / 0.671 & 0.785 / 0.775 & \textbf{0.545} / \textbf{0.540} & 0.664 / 0.697 & 7.1 / 3.4 \\
\midrule
\textbf{\SAS{} (ours)} & \textbf{0.837} / \textbf{0.817} & 0.901 / 0.903 & \textbf{0.745} / \underline{0.686} & \textbf{0.864} / \textbf{0.861} & 4.2 / 2.4 & 0.680 / \underline{0.703} & \textbf{0.798} / \textbf{0.801} & 0.363 / 0.419 & 0.880 / \underline{0.889} & 15.3 / 1.3 \\\bottomrule
\end{tabular*}
\end{table*}

\subsection{Simulation results}
Table~\ref{tab:results} reports TQ and its three channels per controller in simulation under
the ground-truth and the onboard map, and the percentage of obstacle distance the \SAS{}
override held each controller. The manual rides remain the reference: manual control
scores $0.89$, $0.05$ above the best controller, and the difference lies almost entirely in
the clearance channel ($0.87$ against $0.61$--$0.75$ for the controllers), while shock and tilt are within $0.03$
of the operator for the best controllers. Every controller loses quality under the onboard
map, by $0.008$ to $0.069$. \SAS{} loses $0.020$, through its LiDAR-ICP motion cue alone.
Most controllers required the override for 7\% to 15\% of the obstacle distance under
either map to finish the obstacles. The shock channel hardly separates the controllers,
$0.86$ to $0.92$: all of them experienced a similar acceleration, dominated by the rigid
contacts of the simulator rather than by the flipper motion. C-TRAC degrades the most under the onboard map, by $0.069$, and shows the highest
sensitivity to the elevated noise of the onboard map: never overridden under the
ground-truth map, it is held for 7.5\% of the distance under the onboard map. For Yuan, Oehler, and
Mitriakov, the override share dropped under the onboard map: the added noise made them move their flippers, which resulted in successful traversals.

\subsection{Real-robot results}
On the real pallet (Table~\ref{tab:results}), the operator remains the reference: both manual
rides score $0.81$, above every controller. Three findings stand out. \textbf{First}, the best
controllers are Yuan under the onboard map ($0.72$) and Oehler and Mitriakov under the
ground-truth map (both $0.71$). \textbf{Second}, \SAS{} is among the best controllers on the real
robot as well, second under the onboard map behind Yuan. \textbf{Third}, stalls are rarer than in simulation: the map-based controllers that stall
are released by the \SAS{} override for at most 7.1\% of the crossing distance, and, unlike in simulation, the onboard map lowers that share for most of them; the pallet is also easier to
mount on the real robot, whose track flaps catch the edge. The shock channel again has a small spread, $0.76$ to
$0.80$, so the controllers experienced a similar acceleration on the pallet as well. The two map
conditions separate the controllers far less here than in simulation because one crossing is short and
the perception stack is reset before each of them, so the onboard map has little time to drift.

\subsection{Hidden obstacle}
\begin{figure*}[t]
  \centering
  \includegraphics[width=\textwidth]{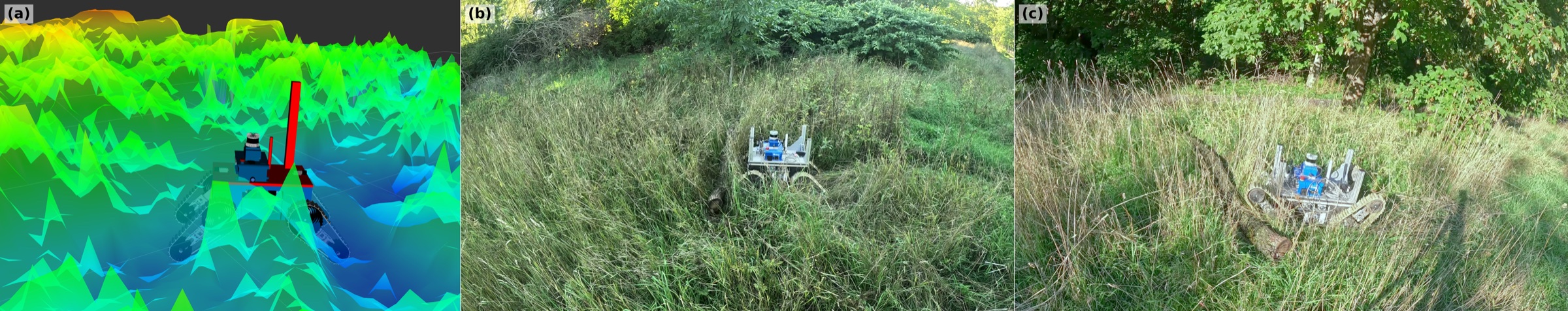}
  \caption{\textbf{Traversal of an obstacle hidden in tall grass by the map-based
  controller Yuan~\cite{yuan2019} and by \SAS.} (a) The onboard map built from the LiDAR point
  cloud during Yuan's crossing: the LiDAR does not penetrate the grass, and so the map shows only the canopy while the log remains obscured. (b) Yuan interprets the gap created by the log in the canopy as a drop and lowers the front flippers to descend; the lowered flippers slide under the log, causing the robot to get stuck underneath it. (c) \SAS{} does not rely on the LiDAR-derived map; therefore, it keeps its default posture and easily crosses the log.}
  \label{fig:grass}
\end{figure*}

Figure~\ref{fig:grass} shows a real experiment: a crossing of a log hidden in tall grass, with Yuan and \SAS. The LiDAR does not penetrate the grass, which causes the onboard map to show the grass
canopy, and the log is absent from it. Yuan interprets the gap that the log creates in the grass as a drop and lowers the front
flippers to descend; the lowered flippers slide under the log, causing the robot to get stuck underneath it. \SAS{} does not rely on the map, keeps its default posture, and crosses the log.

\subsection{Onboard-map error}
\begin{figure}[t]
  \centering
  \includegraphics[width=\linewidth]{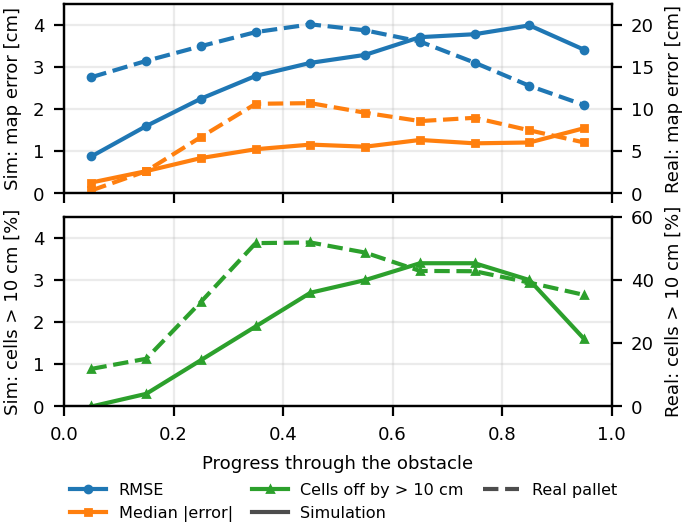}
  \caption{\textbf{Onboard-map error over an obstacle in simulation and on the real pallet.}
  Median over all rides against the progress through the obstacle. Top: the height difference between
  the onboard and the ground-truth map over the body footprint. Bottom: the share of cells
  off by more than \SI{10}{\centi\metre}. Both grow while the robot ascends the obstacle and
  fall again on the descent.}
  \label{fig:maperr}
\end{figure}

Figure~\ref{fig:maperr} shows how the onboard map degrades over an obstacle, in simulation
and on the real pallet. The onboard map is compared with the ground-truth map over the
body footprint (the chassis and \SI{1.5}{\metre} ahead, at a grid resolution of
\SI{5}{\centi\metre}) at every
map update, every \SI{0.25}{\second}. The error grows while the robot ascends the obstacle,
where the ground under the body was last seen from the flat approach, and falls again on
the descent, where the ground ahead comes back into view. On the real robot, the error peaks five
times higher than in simulation, and the share of cells off by more than \SI{10}{\centi\metre}
is fifteen times higher, so the onboard map is not merely shifted but far noisier.

\section{Conclusion}\label{sec:conclusion}
State-of-the-art flipper controllers depend on an elevation map of the occluded ground under
the robot chassis, which drifts with odometry on exactly the terrain where flippers are needed.
The comparison shows which part of a traversal the choice of controller actually
governs. Shock does not separate the controllers: over all thirteen of them, it spans $0.865$ to $0.918$ in simulation and $0.765$ to $0.798$ on the real pallet. Every controller
transmits a similar acceleration to the body, set by the platform and the terrain rather
than by the flipper motion.
What separates them is how they stabilize the body: clearance and
tilt; on the real robot, clearance is the most significant factor, which spreads from $0.301$ to $0.545$. That channel is also the one the transfer from
simulation damages most: every controller drops from $0.68$--$0.75$ in simulation to
$0.30$--$0.55$ on the real robot, while shock and
tilt keep their level. The onboard
map costs quality throughout: in simulation, every controller scores lower under it than
under the ground-truth map, by $0.008$ to $0.069$, against $0.020$ for \SAS{} through its
motion cue alone.

\SAS{} reaches the highest traversal quality of all autonomous controllers in simulation under both
maps, and on the real pallet, it is second under the onboard
map and among the leading controllers under the ground-truth map. It recovers from every stall by its own anti-stuck term. Its weakest channel is on the real pallet, where it holds a clearance of $0.363$ under the
ground-truth map and $0.419$ under the onboard map, below the best map-based controllers because,
 without a terrain preview, it can only raise the chassis once the obstacle is already
under it. 

On a real log hidden in tall grass, invisible to the LiDAR and absent from the
onboard map, the map-based controller Yuan~\cite{yuan2019} got stuck under the log while \SAS{} crossed it.
Two principles as simple as \emph{hold the body level} and \emph{dig out when motion
stops} can reach the traversal quality of the map-based state of the art.

All comparisons reset the perception stack at each obstacle, which limits the drift the
onboard map can accumulate; therefore, the reported map degradation is a lower bound, and a long,
uninterrupted traversal remains to be tested.
Future work targets a more robust motion cue by fusing LiDAR-ICP odometry with camera
odometry, which should hold up in occluded, feature-poor, or flat places where ICP alone struggles, and a fallback scheme that defers to a map-based
controller when the onboard map is trustworthy and to \SAS{} when it is not.

\bibliographystyle{IEEEtran}
\bibliography{root}

\end{document}